\documentclass[journal, letterpaper]{IEEEtran}

\usepackage{graphicx}

\usepackage{url}

\usepackage{amsmath}

\usepackage{textgreek}	
\usepackage{listings}
\usepackage{csvsimple}
\usepackage{longtable}
\usepackage{graphicx}

\usepackage{biblatex}

\begin{document}

	\title{Generation of Realistic Synthetic Raw Radar Data for Automated Driving Applications using Generative Adversarial Networks}
	\author{
    \IEEEauthorblockN{
        Eduardo C. Fidelis\IEEEauthorrefmark{1},
        Fabio Reway\IEEEauthorrefmark{2,4}, 
        Herick Y. S. Ribeiro\IEEEauthorrefmark{1}, 
        Pietro L. Campos\IEEEauthorrefmark{1},     
        Werner Huber\IEEEauthorrefmark{2}, 
        Christian Icking\IEEEauthorrefmark{4},
        Lester A. Faria\IEEEauthorrefmark{1}, 
        Torsten Schön\IEEEauthorrefmark{3}}
        
    \IEEEauthorblockA{\IEEEauthorrefmark{1}Facens University, Brazil\\
    \IEEEauthorrefmark{2}CARISSMA, Technische Hochschule Ingolstadt, Germany\\
    \IEEEauthorrefmark{3}AImotion Bavaria, Technische Hochschule Ingolstadt, Germany\\
    \IEEEauthorrefmark{4}FernUniversität in Hagen, Germany\\
    }%
    \thanks{Corresponding author: Eduardo Fidelis (Eduardo.Fidelis@facens.br).\protect}
    \thanks{GitHub: eduardo-candioto-fidelis/raw-radar-data-generation}%
}
	\maketitle

\begin{abstract}
The main approaches for simulating FMCW radar are based on ray tracing, which is usually computationally intensive and do not account for background noise.
This work proposes a faster method for FMCW radar simulation capable of generating synthetic raw radar data using generative adversarial networks (GAN).
The code and pre-trained weights are open-source and available on GitHub.
This method generates 16 simultaneous chirps, which allows the generated data to be used for the further development of algorithms for processing radar data (filtering and clustering).
This can increase the potential for data augmentation, e.g., by generating data in non-existent or safety-critical scenarios that are not reproducible in real life.
In this work, the GAN was trained with radar measurements of a motorcycle and used to generate synthetic raw radar data of a motorcycle traveling in a straight line.
For generating this data, the distance of the motorcycle and Gaussian noise are used as input to the neural network.
The synthetic generated radar chirps were evaluated using the Frechet Inception Distance (FID).
Then, the Range-Azimuth (RA) map is calculated twice: (1\textsuperscript{st}) based on synthetic data using this GAN and (2\textsuperscript{nd}) based on real data. 
Based on these RA maps, an algorithm with adaptive threshold and edge detection is used for object detection.
The results have shown that the data is realistic in terms of coherent radar reflections of the motorcycle and background noise based on the comparison of chirps, the RA maps and the object detection results.
Thus, the proposed method in this work has shown to minimize the simulation-to-reality gap for the generation of radar data.
\end{abstract}

\section{Introduction}
	
	Automated driving relies on different environment sensors, such as camera, radar and lidar, for object and obstacle detection in the surroundings of the ego-vehicle.
This task must be performed with high accuracy so that the automated vehicle can safely create a driving plan and execute a collision-free trajectory.
Radar sensors are capable of measuring the relative position and velocity of other road users with high accuracy.
They are also the most robust type of sensor that can operate in adverse weather conditions such as rain and fog, as their performance does not degrade as much as cameras and lidars \cite{Hasirlioglu.2020, Hasirlioglu.2017, Yoneda.2019, Reway.2018, FILGUEIRA.2017}.
For this reason, radar sensors are widely used by the automotive industry, which already integrates them into production vehicles for the operation of SAE Level 1, 2 and 3 systems such as adaptive cruise control (ACC), automatic emergency braking (AEB) and traffic jam pilot. 
However, the safe operation of automated driving systems (ADS), as defined for SAE Level 4 and 5, may rely on further developments based on radar sensors to expand the operational design domain (ODD) of such systems.

\begin{figure}[t!]
        \centering 
        \includegraphics[width=\columnwidth]{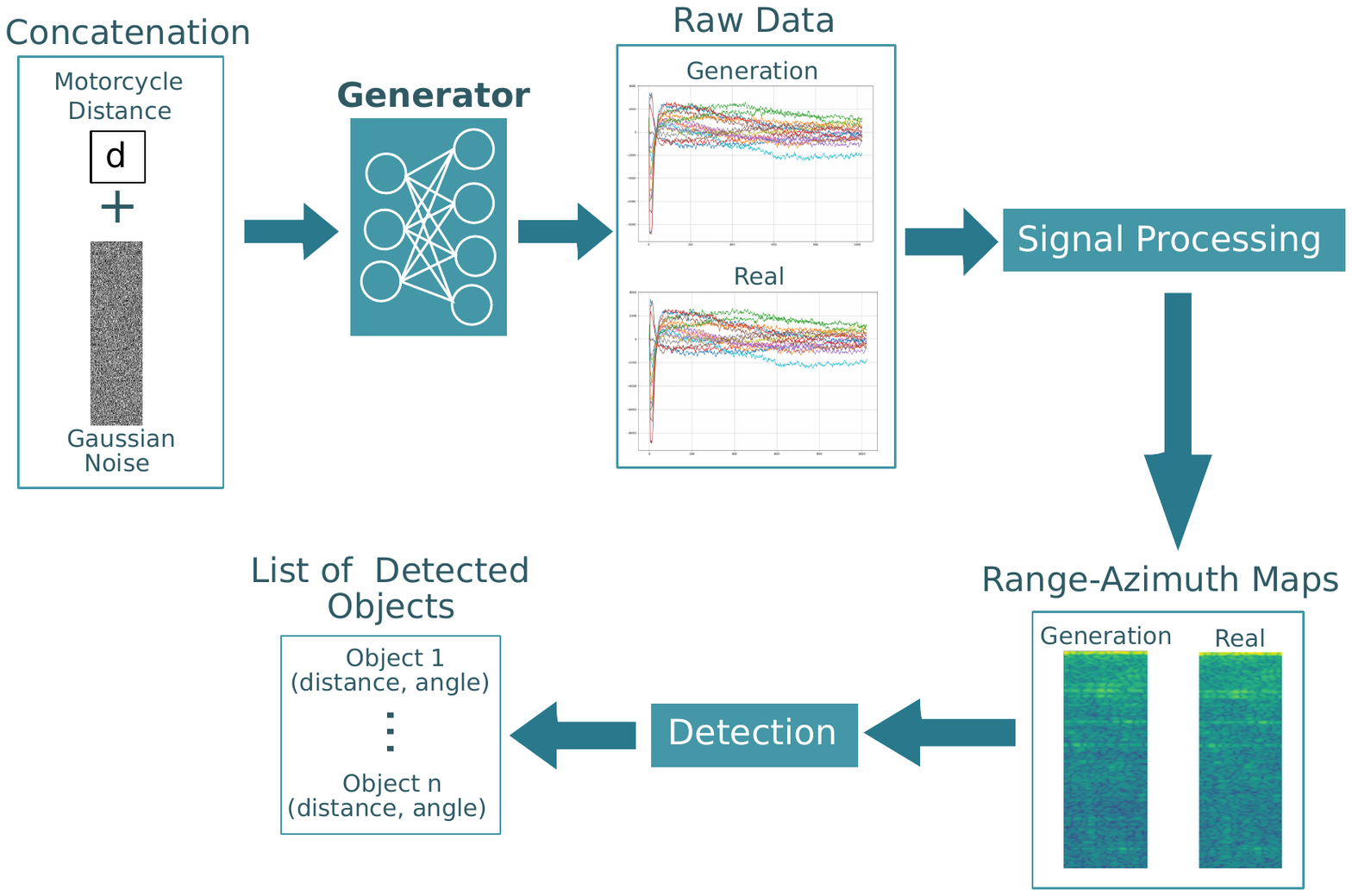}
        \caption{This shows the process from generation to target detection. This work creates an alternative that is less computationally expansive than ray tracing.}
        \label{fig:teaser}
\end{figure}

Simulation has become increasingly important in the development of automated driving.
The German Regulation for the operation of motor vehicles with automated functions enables verification and validation (V\&V) methods using simulation \cite{DIPVeror95:online}.
It emphasizes that there should be no significant discrepancy between test results obtained in the virtual and real domains, which can also be defined as simulation-to-reality gap \cite{Reway.2020, Wachtel.2021, Ngo.2021}.
The use of simulation is already established, for example, for the generation of synthetic image data for data augmentation and training of neural networks.
However, it is critical that sensor models are able to produce realistic measurements that have similar characteristics to those obtained in the real environment, i.e., that the simulation-to-reality gap is minimum. 
Thus, simulation-based development of further sensor data processing algorithms can become more iterative by reducing the effort required to collect data in the real world for different scenarios and test conditions. 


Different types of sensor models were developed to allow the generation of synthetic radar measurements: ideal, physical and artificial intelligence-based sensor models \cite{Ngo.2021}.
Ideal sensor models are capable of producing "ground-truth" data that are error-free or where noise has been added to the ideal measurements.
Objects are always visible based on their position, and the range and field-of-view (FOV) of the parameterized sensor model.
Physical sensor models rely on ray-tracing or ray-casting methods to reproduce radar wave propagation in a virtual environment.
This involves generating rays for each radar transceiver that reflect off the geometry of a 3D object to create a radar reflection point.
Therefore, the computational cost of using such a sensor model can be high, depending on the required number of points to be generated.
An artificial intelligence-based sensor model (or data-driven model) can be a fair compromise between realism and performance.
These are able to reproduce measurements from a specific sensor component, based on previous recorded measurements acquired with the radar sensor.

The contribution of this work is the conceptualization and implementation of a sensor model based on Generative Adversarial Networks (GAN) for the generation of raw radar data.
This model has been trained with data acquired with a frequency modulated continuous wave (FMCW) radar and is capable of generating multiple intermediate frequency (IF) signals for a target (of class motorcycle) based on its range.
Figure \ref{fig:teaser} shows the summary of the approach adopted in this work.
The trained weights and code are available on GitHub\footnote{github.com/eduardo-candioto-fidelis/raw-radar-data-generation}.

\section{Related Works}

This section introduces the FMCW radar and summarizes the state-of-art methods for the generation of synthetic radar data.

FMCW radars transmit a signal (TX signal) composed by chirps whose frequency is modulated to measure the position and velocity of targets. 
The targets in front of the radar reflect the emitted TX signal, which are then received back by the antennas of the radar. 
The reflected portion of this signal is then defined as RX signal. 
In the next step, the TX and RX signals are mixed, resulting in an IF signal that can be further processed with FFTs.
The FFT operation can be used to generate the Range-Doppler (RD) map, based on processing multiple sequential chirps, and the Range-Azimuth (RA) map, based on processing multiple simultaneous chirps.
From the RD map the velocity of the corresponding targets can be estimated, and from the RA map the position is estimated \cite{fundamentals}.



The authors in \cite{launching} developed a physical radar model for generating synthetic data as for the IF signal based on ray launching.
Their proposed model seeks for paths between the transmitter antenna and the target in the virtual environment, considering the reflections in other obstacles in the scene. 
This enables the simulation of the physical propagation of a radio electromagnetic wave in a given environment. 
This is used to generate the IF signal, being is discriminated for time and intensity by a simulated A/D converter.
Another approach using ray tracing is the work proposed in \cite{fourier}.
In this work, the authors use the Fourier Tracing technique that does not require expensive FFT computations, and it is designed to capture main features of the radar measurements, such as interference and multi-path reflections.
This increases realism in simulation since interference and multi-path reflections are sources of radar measurements imprecision. 
Other works also focus on radar simulation, such as \cite{mimo, 8835755, geosciences10010009, 9472519}.
However, the disadvantage of such approaches is the high computational effort and the time required to run the simulation, which makes the real-time execution of this model challenging  \cite{deep, marcio}.

Other approaches to radar modelling based on machine learning could facilitate the generation of radar data.
In \cite{deep}, an architecture is proposed to generate RD maps using an object list and a grid that specifies the environment and its materials proprieties, to capture the interaction of the radar signals with the environment.
The authors achieved the most realistic results for RD maps generation using a mixture of GAN and Variational Autoencoder (VAE), called VAE-GAN.
The authors in \cite{marcio} propose the creation of RD maps based on a GAN and Deep Convolutional Autoencoders (CAE).
The purpose of the radar data generation is the data augmentation for training an object detection on RD maps.
An AE is used to reduce the noise in the RD maps, highlighting the target. 
These RD maps are used to train a GAN conditioned by velocity and distance. 
Then another AE is used to predict the next four consecutive RD maps, and noise is added to these generations.
The generations are used to train the detection algorithm, which is then evaluated against real data.

The work in \cite{ltr} uses GAN to convert a LiDAR point cloud into sparse pseudo-image (RA maps). 
For achieving that, an architecture called L2R is proposed. 
This means that the representation of measured data can be transferred to another different domain. 
Then, sensor data is collected with a car equipped with a LiDAR and a radar sensors.
This dataset is used to train the GAN that is able to generate radar data based on LiDAR point clouds.


This work proposes the generation of IF signal using GAN.
The proposed model is capable of generating raw data for an specific object class (motorcycle) given a determined distance. Which requires less computation compared with the ray tracing methods.

\section{Material and Methods}

For data collection, a total of 37 tests were conducted with the INRAS Radarlog while the motorcycle was in motion.
The real IF signal captured by the radar is used for training.
For simplification purposes, the data of only one experiment was used to train the GAN model.
This experiment consists of the motorcycle moving in a straight line towards the radar. 


GAN is a neural network architecture for generating realistic samples similar to those in the training set.
It consists of two models, a discriminator ($D$) and a generator ($G$).
$G$ receives noise as input and outputs some samples, $D$ compares how much these samples differ from the real ones.
Then, $D$ transfers this difference to $G$ via back-propagation and improves them to produce even better generations \cite{gan}. 
In this work, the WGAN-GP was used because it is expected to be more stable than the classic GAN architecture \cite{wgangp}.

This work focuses on the generation of 16 simultaneous chirps per time of a motorcycle at a given distance.
To achieve that, the WaveGAN was used. 
Figure \ref{fig:teaser} depicts this approach.
The architecture uses one-dimensional convolutions instead of the usual two-dimensional ones since the data is one-dimensional with multiple channels.

\begin{figure}[]
        \centering 
        \includegraphics[width=\columnwidth]{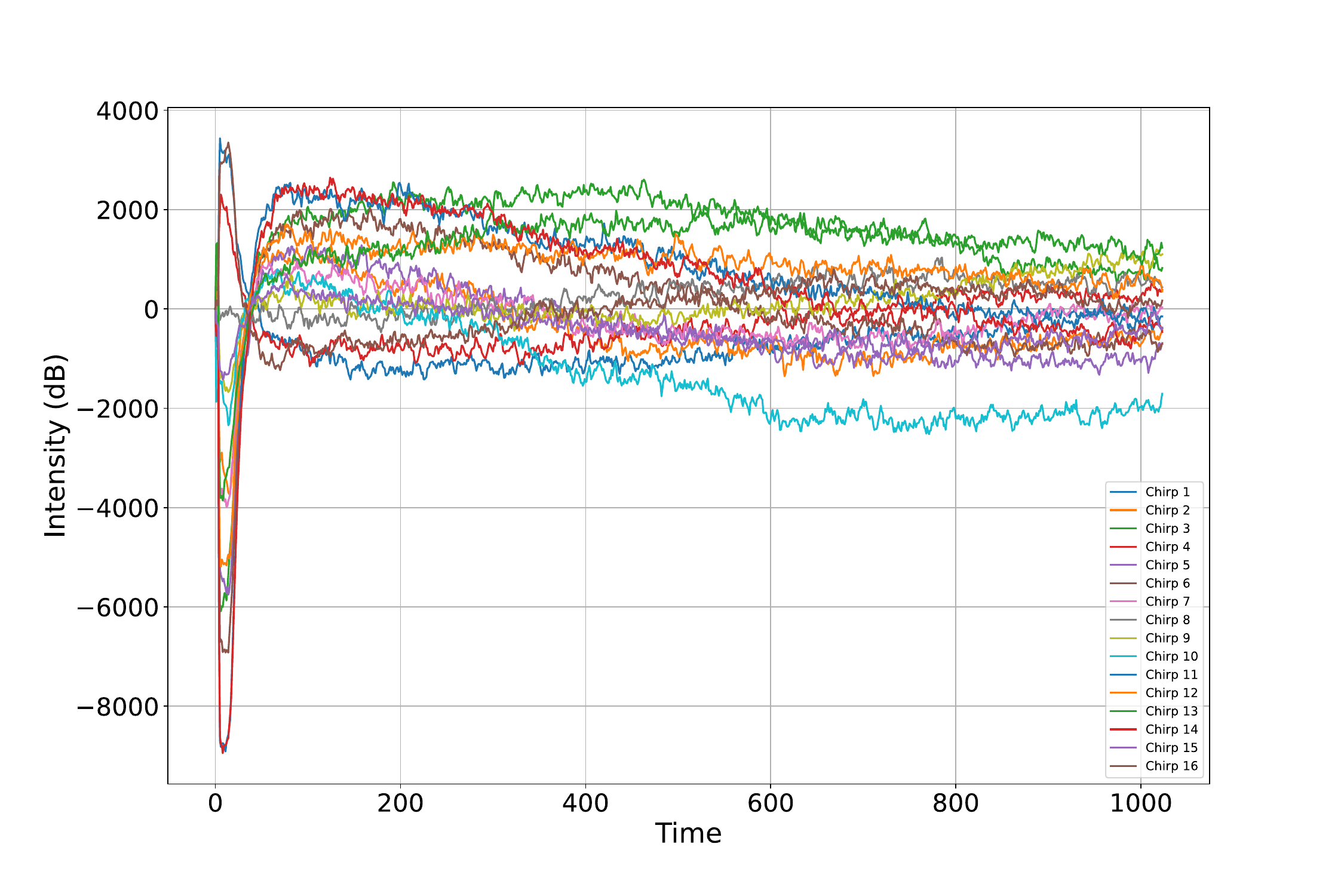}
        \caption{Real bunch of chirps -- each color represents one of the 16 chirps.}
        \label{fig:chirp}
\end{figure}

\begin{figure}[b!]
        \centering 
        \includegraphics[width=0.7\columnwidth]{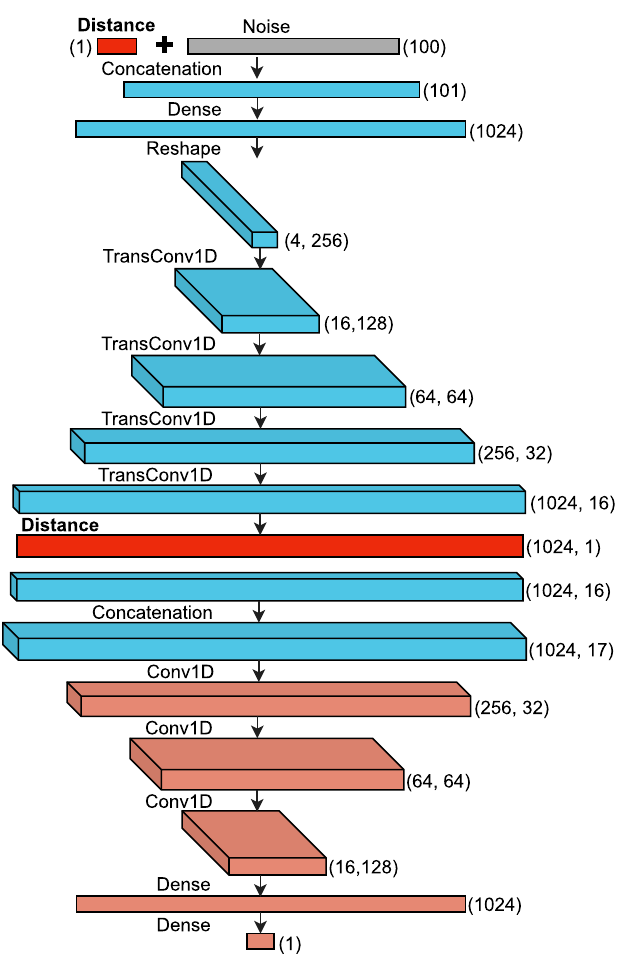}
        \caption{Our proposed architecture: generator parts are blue and discriminator parts are red. ReLU activation function is used after the conv in the generator and the LeakyReLU with lambda 0.2 in the generator.}
        \label{fig:arch}
\end{figure}

An example of a bunch of chirps acquired in a real experiment is shown in the Figure \ref{fig:chirp}.
The first 250 values of each chirp are not used in the process to calculate the RD and RA maps, so they were set to zero not to interfere in the quality of the synthetic generations. 
The used architecture used is a variation of the WaveGAN, having some filter values and number of layers adjusted; the final model is shown in Figure \ref{fig:arch}. 
For conditioning the generations, the distance value is concatenated with the noise that was used as input to the $G$ and as an additional feature to the $D$. 
Since $G$ uses the activation function $tanh$ whose output ranges from -1 to 1, the IF signal data and the distance values for $D$ were scaled to the same range.
The distance values for the $G$ were scaled with the $standard scaler$ to be closer with the scale of the Gaussian noise, which showed to reduce the training convergence time. 
The model was trained for 4655 epochs -- and it was observed that a greater epoch number does not seem to improve the quality of the generations.


In order to evaluate the synthetic data generated by GANS, the metric Frechet Inception Distance (FID) can be used \cite{fid}. 
It uses a feature extractor that is a Convolutional Neural Network (CNN) trained for supervised learning.
Using FID, the fully connected layers are removed at the top and the output distribution of the convolutions for the real and generations are compared.
The closer the FID value of the synthetic data is to the FID value of the real data, the better the generations.

\begin{figure}[t!]
        \centering 
        \includegraphics[width=0.7\columnwidth]{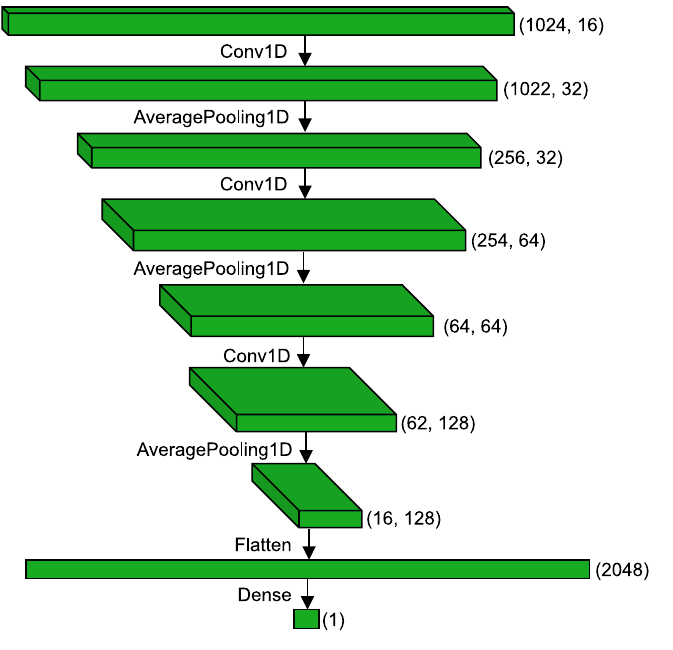}
        \caption{Regression model architecture. ReLU is used in all convolution layers and L2 regularization with lambda of 0.5 on the conv and dense layers.}
        \label{fig:regressor}
\end{figure}

Another metric was used to check if the generations are new and not copied from the training data.
This verify whether the samples of the generations are, on average, as far away from the training data as the training data itself.
Euclidean distance is used to measure this distance, a technique also known as nearest neighbor (NN).

\section{Results}

\begin{figure}[h]
        \centering 
        \includegraphics[width=0.2\textwidth]{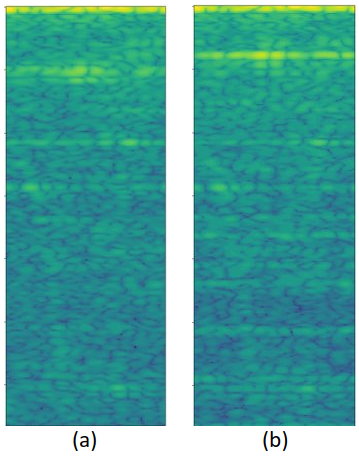}
        \caption{Sample of real RA map (a) and of generated RA map (b).}
        \label{fig:rareal}
\end{figure}

The results of the evaluation metrics indicate that $G$ is able to generate background noise, which can be observed in the images on the regions of less intensity.
Figure \ref{fig:rareal} shows a comparison between a real and a generated RA map using the proposed model.

In this work, a motorcycle distance prediction model was used as a feature extractor -- and it achieved a mean absolute error of 2.69 meters, the range of possible values is between 0 and 25 meters. 
The architecture is presented in Figure \ref{fig:regressor}.

In the Table \ref{tab:metrics}, the quantitative metrics and their corresponding results are presented. 
The analysis of the FID score shows that the generations are close to the real data, once theses scores for generation-test are lower than between training-test.
Also, the NN value of the comparison generation-training is higher than training-training, but not significantly lower than training-test, what suggest that the generations are not copies of the training set, but still rely on it.
Comparing the motorcycle detected distance in generated samples with the ones requested to the $G$, they differ in average 1.4 m. However this distance already exists in the training set.
These results indicate that this method is able to generate new background noise, which is caused by interactions of the radar signal with the environment.

In the experiments, the generator took 3.92 seconds to generate 6000 chirps on a computer with 16 GB RAM and a Core i7-10750H, which gives an average time sample generation of 0.65 ms.

\begin{table}[]
\centering
\caption{FID id the FID score and NN is the Nearest Neighbour value, which are normalized by the first row.}
\label{tab:metrics}
\begin{tabular}{l|cc|c}
\hline
           & \multicolumn{2}{c|}{FID}              & NN       \\ \cline{2-4} 
           & \multicolumn{1}{c|}{Training} & Test  & Training \\ \hline
Training   & \multicolumn{1}{c|}{0.06}     & 18.18 & 1.00     \\
Test       & \multicolumn{1}{c|}{18.18}    & 0.01  & 2.83     \\
\textbf{Generation} & \multicolumn{1}{c|}{\textbf{0.51}}     & \textbf{18.12} & \textbf{2.37}     \\ \hline
\end{tabular}%
\end{table}

\section{Conclusion}

FMCW radars are widely used in the automotive industry to achieve SAE Levels 2 and beyond for automated driving.
In this work, a method for generating raw radar data based on GAN is proposed.
This method can be used to generate a residual of the IF signal, which can be used to facilitate the development of further algorithms for radar data processing, such as data filters, clustering, and object detection and classification.
This method has the advantage of requiring less computation and time compared to ray tracing methods. 
Considering this, the proposed model could be implemented in simulators -- such as CARLA \cite{Carla.2017} -- and could be an alternative to ray tracing approaches.

Future work can explore extrapolating the target position using environmental information in the scene to simulate the interaction of the radar signal with that scene. 
Also, the generation of multiple consecutive chirps could be explored to obtain an RD map and determine the velocity of the targets. 

\section*{Acknowledgment}
\addcontentsline{toc}{section}{Acknowledgment}
\scriptsize
We gratefully acknowledge the support provided by Fundação de Desenvolvimento da Pesquisa – Fundep Rota 2030/Linha V [27192*09], Facens University and Technische Hochschule Ingolstadt

\printbibliography

\end{document}